\documentclass[runningheads]{llncs}

\usepackage[mobile]{eccv}

\usepackage{eccvabbrv}

\usepackage{graphicx}
\usepackage{booktabs}

\usepackage[accsupp]{axessibility}  %

\usepackage[pagebackref,breaklinks,colorlinks]{hyperref}
\usepackage{orcidlink}

\usepackage{wrapfig}

\newcommand{\myparagraph}[1]{\vspace{1pt}\noindent{\bf{#1}}}
\newcommand{\benchmarkName}{\texttt{EgoCVR }}
\newcommand{\benchmarkNameNS}{\texttt{EgoCVR}}
\newcommand{\methodName}{\texttt{TFR-CVR }}
\newcommand{\methodNameNS}{\texttt{TFR-CVR}}

\newcommand{\xmark}{{\color{myRed}\ding{55}}}%
\newcommand{\cmark}{{\color{myGreen}\ding{51}}}
\usepackage{xcolor}
\usepackage{pifont} 
\definecolor{myGray}{rgb}{0.5, 0.5, 0.5}
\definecolor{myRed}{rgb}{0.808,0.067,0.149}
\definecolor{myGreen}{rgb}{0.067,0.708,0.149}
\usepackage{multirow}
\newcommand{\xmarkg}{{\color{myGray}\ding{55}}}%

\usepackage{longtable}
\usepackage[skins]{tcolorbox} 
\tcbuselibrary{breakable}
\usepackage{float}
\newfloat{Prompt}{htbp}{loa}
\usepackage{placeins}

\newcommand{\rulesep}{\unskip\ \vrule\ }

\begin{document}

\title{EgoCVR: An Egocentric Benchmark for Fine-Grained Composed Video Retrieval} 

\titlerunning{EgoCVR}

\newcommand*\samethanks[1][\value{footnote}]{\footnotemark[#1]}
\author{Thomas Hummel\thanks{Denotes equal contribution}\inst{1,2}\orcidlink{0000-0003-3201-360X},
Shyamgopal Karthik\samethanks\inst{1,2}\orcidlink{0009-0006-5554-1375},\\
Mariana-Iuliana Georgescu\inst{2}, Zeynep Akata\inst{2,3}\orcidlink{0000-0002-1432-7747}}

\authorrunning{T. Hummel et al.}

\institute{$^1$Tübingen AI Center, University of Tübingen 
 \\ $^2$Helmholtz Munich, MCML 
$^3$TU Munich\\
\email{\{thomas.hummel, shyamgopal.karthik\}@uni-tuebingen.de}}

\newcommand{\blue}[1]{\textcolor{blue}{#1}}
\newcommand{\green}[1]{\textcolor{green}{#1}}
\newcommand{\gray}[1]{\textcolor{gray}{#1}}

\maketitle

\begin{abstract}
In Composed Video Retrieval, a video and a textual description which modifies the video content are provided as inputs to the model. The aim is to retrieve the relevant video with the modified content from a database of videos. 
In this challenging task, the first step is to acquire large-scale training datasets and collect high-quality benchmarks for evaluation. In this work, we introduce \benchmarkNameNS, a new evaluation benchmark for fine-grained Composed Video Retrieval using large-scale egocentric video datasets.  \benchmarkName consists of 2,295 queries that specifically focus on high-quality temporal video understanding. We find that existing Composed Video Retrieval frameworks do not achieve the necessary high-quality temporal video understanding for this task. %
To address this shortcoming, we adapt a simple training-free method, propose a generic re-ranking framework for Composed Video Retrieval, and demonstrate that this achieves strong results on \benchmarkNameNS. Our code and benchmark are freely available at \url{https://github.com/ExplainableML/EgoCVR}. %
\end{abstract}

\section{Introduction}

\begin{figure}[t]
    \centering
    \includegraphics[width=\textwidth]{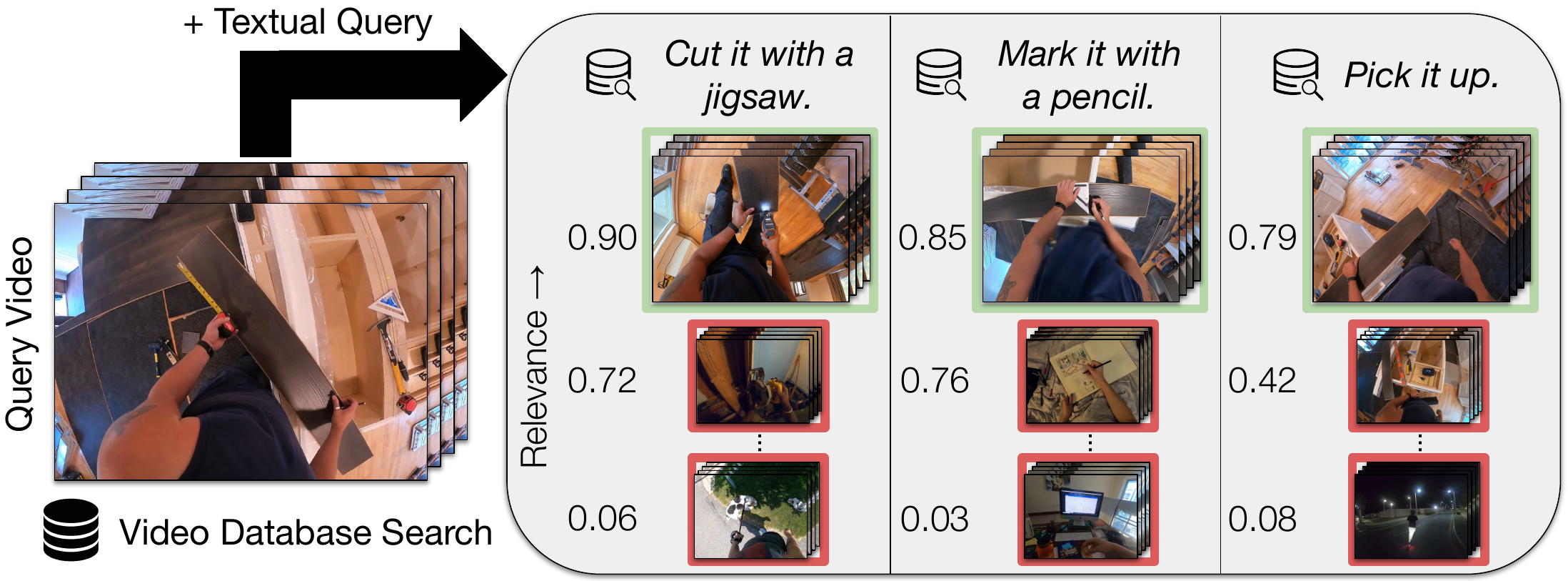}
    \caption{The goal of the Composed Video Retrieval (CVR) task is to retrieve the correct video using both a query video and a textual video modification instruction that describes the semantic changes required from the query video. %
    }
    \label{fig:motivation}
\end{figure}

Recent advances in Vision-Language Models (VLMs) have enabled video search through free-form textual descriptions. However, expressing complex queries, especially those involving subtle transformations or actions, remains challenging with purely text-based searches. 
In the image domain, Composed Image Retrieval (CIR)~\cite{cirr,tirg,guo2019fashion} has emerged as a related task where a user provides a reference image and a textual description of the desired modification. %
In the video domain, the corresponding task is coined as Composed Video Retrieval (CVR)~\cite{ventura2023covr} where the aim is to retrieve videos from a database given a reference video and a textual query that describes how the reference video should be modified. For example, a user searching through a long video might provide a short reference clip showing construction work along with a textual description such as ``\texttt{make the person cut with a jigsaw instead}'' to pinpoint the precise video they are looking for (See Figure~\ref{fig:motivation}).
CVR remains a relatively under-explored area, posing unique challenges due to the added complexity of effectively utilizing the temporal information inherent in videos. CVR is extremely challenging because it requires understanding both the visual and textual inputs and composing them to retrieve the desired video efficiently.

 A major step towards tackling the CVR challenge is the introduction of the large-scale WebVid-CoVR training set and a smaller evaluation benchmark. These datasets are automatically collected by using existing video-text datasets looking for pairs that differ only in a single word in the caption, and using a Large Language Model (LLM)~\cite{llama2} to generate the textual instruction. %
The final training set contains over 1.6M triplets, which is extremely useful for the CVR task. However, the evaluation set quality is quite limited due to the automatic dataset construction. For instance, %
most of the modifications predominantly focus on the color, shape, and adding/removing objects from the scene that do not require temporal understanding (see Figure~\ref{fig:dataset_temporal}). Therefore, the task can be tackled with a single image rather than a video, \eg a vision-language model~\cite{blip} trained on the image level achieves state of the art. %

In this work, we %
propose to create an evaluation set for the Composed Video Retrieval task that requires holistic video understanding to obtain strong performance. To achieve this, we propose \benchmarkNameNS, a manually curated and high-quality evaluation set with 2,295 videos sourced from the Ego4D\cite{ego4d} dataset. Our \benchmarkName dataset consists of a query and target clip sourced from the same long video and the textual modifier asking for a subtle change in the action being performed in the clip. As a result, models need to have strong video understanding to be able to achieve strong performance in our evaluation setting.

Furthermore, we evaluate on our new benchmark several methods designed for cross-modal retrieval, consisting of vision-language models such as CLIP\cite{clip}, BLIP~\cite{blip}, the video-based method LanguageBind~\cite{languagebind}, as well as the egocentric video model EgoVLP~\cite{egovlp,egovlpv2}, by adapting them to the CVR task. %
Naively adapting vision-language models to perform the CVR task, even if the model was finetuned on the large benchmark, does not work well, \eg BLIP$_{\text{CoVR}}$ finetuned for the CVR task on 1.6M triplets performs poorly on \benchmarkNameNS. \\
To address this shortcoming, we propose to adapt a training-free method proposed for Composed Image Retrieval~\cite{cirevl} to the Composed Video Retrieval task. When employed with a generic re-ranking strategy, this approach, which we name \methodNameNS, achieves the best results among all considered methods in various evaluation settings. 
To summarise, we make the following contributions:
\begin{itemize}
    \item We propose \benchmarkNameNS, a benchmark with 2,295 queries, to evaluate vision-language models for the task of Composed Video Retrieval.
    \item We evaluate several vision-language models with varying configurations on our benchmark and find that existing models, even when finetuned for Composed Video Retrieval, have several shortcomings on the action-focused\\\benchmarkName benchmark.
    \item Finally, we demonstrate that our proposed training-free \methodName method, along with a generic re-ranking framework, achieves strong performance on the \benchmarkName benchmark.
\end{itemize}

\section{Related Work}
\myparagraph{Video-Language Models and Retrieval.} Early work on video retrieval often focused on extending retrieval approaches from images to videos by aggregating image features within a video~\cite{dong2018predicting,otani2016learning,torabi2016learning,xu2015jointly}. However, with the introduction of large-scale video-text datasets~\cite{howto100m,msrvtt,vatex,webvid}, and contrastive language-image pre-training~\cite{clip,blip,blip2}, there have been several models proposed for the task of video-text retrieval~\cite{luo2022clip4clip,imagebind,languagebind,bain2022clip}. Due to the growing popularity of egocentric video datasets~\cite{ego4d,egoexo4d}, video-language models have been proposed that specifically focus on this setting~\cite{lavila,egovlp,egovlpv2}. %
However, while there has been growing interest in developing video-based foundation models~\cite{vast,mplug}, these have all been focused on captioning and video-text retrieval. Different from this, we show how existing video-text models can be utilised for fine-grained Composed Video Retrieval.%

\myparagraph{Composed Image Retrieval.} The task of Composed Image Retrieval (CIR) has found significant application in conditional search~\cite{guo2019fashion,fashion200k,tirg}, where users perform interactive dialogue to refine a given query image toward retrieving %
specific items. Classical techniques often employ custom models that project text-image pairs into a common embedding space~\cite{tirg,combiner,chen2020image,chen2020learning,lee2021cosmo,anwaar2021compositional} or use cross-modal attention mechanisms~\cite{artemis}. With the advent of vision-language foundation models~\cite{bommasani2021opportunities,clip,align}, interest in CIR has surged, especially in zero-shot settings that avoid the need for task-specific models. Recent works either attempt to train models that avoid the necessity for paired triplets~\cite{pic2word,searle,sentenceprompts,lincir,tang2023context,chen2023pretrain} or train models on large datasets that then generalise to a wide variety of scenarios~\cite{compodiff,ventura2023covr,levy2023data,transagg}. There have also been several datasets and benchmarks proposed for Composed Image Retrieval including large-scale generic datasets such as CIRR~\cite{cirr}, CIRCO~\cite{searle}, as well as fine-grained evaluation benchmarks focusing on fashion~\cite{fashion200k,guo2019fashion}, fine-grained attributes~\cite{vaze2023genecis}, sketches~\cite{gatti2024composite} or birds~\cite{forbes2019neural}. In this work, we are inspired by both CIR methods~\cite{cirevl,sun2024training} as well as CIR benchmarks~\cite{cirr,vaze2023genecis} in curating a fine-grained Composed Video Retrieval dataset, as well as proposing general methods that can tackle this task.\\
\myparagraph{Composed Video Retrieval.} To the best of our knowledge, the only existing benchmark available for Composed Video Retrieval is WebVid-CoVR~\cite{ventura2023covr}. Further, the only models tailored for it are BLIP models finetuned on the training set of WebVid-CoVR~\cite{ventura2023covr,thawakar2024composed}. Concurrent to our work, the task of \textit{video detours}~\cite{ashutosh2024detours} was introduced, which focused on retrieving and localising temporal segments within long videos using free-form textual queries and the query video, specifically for instructional videos. %
In this work, we propose a fine-grained evaluation benchmark for Composed Video Retrieval with two evaluation settings, along with a training-free method using video-specific models for this task. 

\section{\benchmarkName: An Egocentric Benchmark Dataset for Composed Video Retrieval}
In Section~\ref{sec:cvr}, we first formally define the task of Composed Video Retrieval, while in Section~\ref{sec:dataset}, we describe our dataset construction methodology in detail.
\subsection{Problem Definition}
\label{sec:cvr}
The Composed Video Retrieval task was first introduced by Ventura \etal~\cite{ventura2023covr}. 
Let $\mathcal{V}$ denote the space of videos and $\mathcal{T}$ the space of textual instructions. Given a query video $q_v \in \mathcal{V}$ and textual instruction $q_t \in \mathcal{T}$, the goal of composed video retrieval (CVR) is to identify the modified video $v \in \mathcal{D}$ from a database of videos (gallery) $\mathcal{D} = \{v_1,\dots,v_n\}$, where $n$ is the number of videos in $\mathcal{D}$, that most closely represents the semantic modifications described by $q_t$.
The task can be formalized as a scoring function $\Phi: \mathcal{V}\times\mathcal{T}\times\mathcal{D} \rightarrow {\rm I\!R}$. This function measures the similarity between the query video $q_v$, the modification text $q_t$, and each candidate video $v_i$ in the database, $0 \leq i \leq n$. The video with the highest score according to $\Phi$ is deemed the optimal retrieval result. 

The scoring function $\Phi$ is implemented by representing videos and text within a shared embedding space. We denote the video encoder as $\Psi_v: \mathcal{V} \rightarrow \mathbb{R}^d$ and the text encoder as $\Psi_t: T \rightarrow \mathbb{R}^d$, where $d$ is the dimension of the embedding space. The video encoder $\Psi_v$ processes either single frames (with averaged frame-level embeddings) or frame sequences using a temporal video encoder. The text encoder $\Psi_t$ embeds the modification instructions into the same space as $\Psi_v$. Text and video embeddings are then combined to form a multi-modal video-text embedding $q_{v,t}$ using a fusion function $\Psi_q: \{q_v, q_t\} \rightarrow \mathbb{R}^d$. Candidate videos from the database $\mathcal{D}$ are also encoded using $\Psi_v$. Finally, the cosine similarity is used as a matching score between the query embedding $q_{v,t}$ and each candidate video embedding $v_i$, $v_i \in \mathcal{D}, 0  \leq i \leq n$.

\subsection{From Egocentric Videos to Composed Video Retrieval} \label{sec:dataset}

We collect videos and the corresponding annotations with the narrations (in free-form text) from the Ego4D Forecasting Hand and Object (FHO) task\footnote{The task can be viewed here: \url{https://ego4d-data.org/docs/tutorials/FHO_Overview/}}. As this task focuses on understanding and anticipating human-object interactions, we ensure that collected videos contain frequent and diverse interactions with clear visual quality and a broad range of everyday objects. The FHO task provides short video narrations describing actions and object interactions. An example of a narration is ``\texttt{\#C C trims the blue cardboard to a circular shape with the scissors in her right hand.}''. 
\begin{figure}[t]
    \centering
    \includegraphics[width=0.8\textwidth]{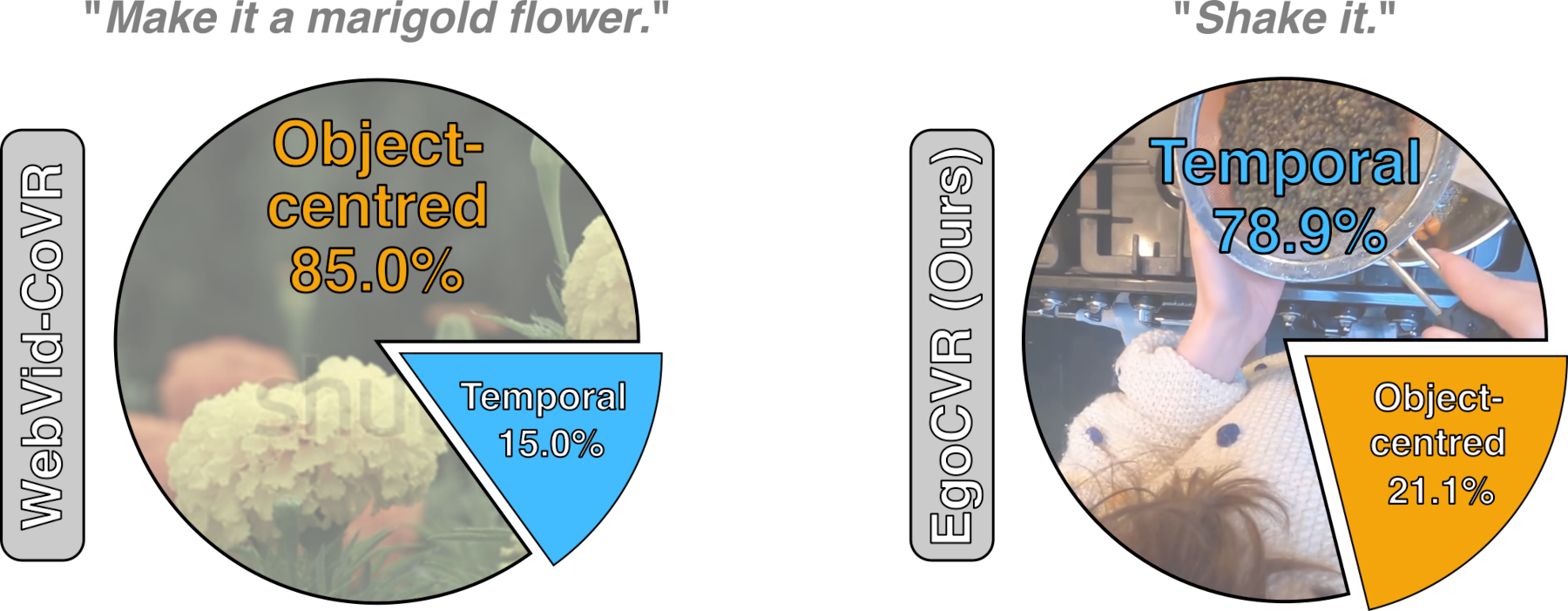} %
    \caption{\benchmarkName focuses to a significantly greater extent on temporal and action-related modifications (blue) as opposed to object-centred modifications (orange) when compared to the previously existing WebVid-CoVR-Test benchmark~\cite{ventura2023covr}.} 
    \label{fig:dataset_temporal}
\vspace{-5pt}
\end{figure}

The dataset includes 155k narrations, each associated with 2-8 second video clips extracted from 1,250 long-form videos. 
We reduce the 155k densely annotated clips to 9k distinct clips by automatically filtering out clips with temporal overlap, ensuring a higher likelihood of single, focused actions within each clip.
Our dataset annotation process aims to find pairs of videos that have subtle differences. While previous work~\cite{ventura2023covr} applied an automated video-matching process by searching for single-word differences in video captions, \benchmarkName is created using a careful manual annotation process outlined in the following. 

We manually search for possible video pairs within long videos, \ie video pairs originate from the same source video. Creating annotations from the same source video allows for fine-grained comparisons where the primary difference between video pairs is the controlled textual modification.  %
We identify matching pairs through similarity in their narrations, \ie the narrations differ in a single semantic concept like actions (\eg rinsing vs. rubbing) or objects (\eg knife vs. spoon). During annotation, an emphasis was put on creating pairs where temporal modifications are prioritised. 
We manually disregard annotation pairs through visual inspection when i) the narrations do not accurately describe the clip, ii) the narrated actions or objects are visible only for a fraction of the clip, or iii) the presence of multiple actions would result in ambiguous samples. 
When multiple videos with the same narration are present (\ie ``\texttt{\#C C puts down the piece of cloth.}'' and ``\texttt{\#C C puts down the cloth.}''), we group the clips together. This allows us to create samples in \benchmarkName with multiple ground truth targets, even for narrations that do not perfectly match with an exact textual search.
This annotation process resulted in a total of 2,295 queries with an average of 1.2 ground-truth targets per query.

\myparagraph{Creating Textual Video Modification Instructions.}
We create textual video modification instructions from the video narrations of paired clips. 
Ideal modification instructions clearly describe the most prominent change that needs to be applied to the query video to get to the desired target video. 
We design these modifications to be as concise as possible while still conveying all the relevant information. %
Instructions in \benchmarkName provide only the minimum necessary semantic difference between the query and the target videos. For instance, the instruction ``\texttt{Rinse it instead.}'' does not provide information on which object to rinse. To create the text modifications, we utilise the reasoning capabilities of LLMs to generate concise instructions that describe the transformation from the provided query clip narration to the target clip narration. As LLM, we employ GPT-4~\cite{gpt4} and provide the LLM with a list of 15 in-context examples~\cite{dong2023survey} together with a clear instruction prompt (more details in the supplementary). We illustrate examples from \benchmarkNameNS, as well as how it contrasts with typical samples from WebVid-CoVR in Figure~\ref{fig:dataset_qualitative}.\\
\begin{figure}[t!]
    \centering
    \includegraphics[width=1.0\textwidth]{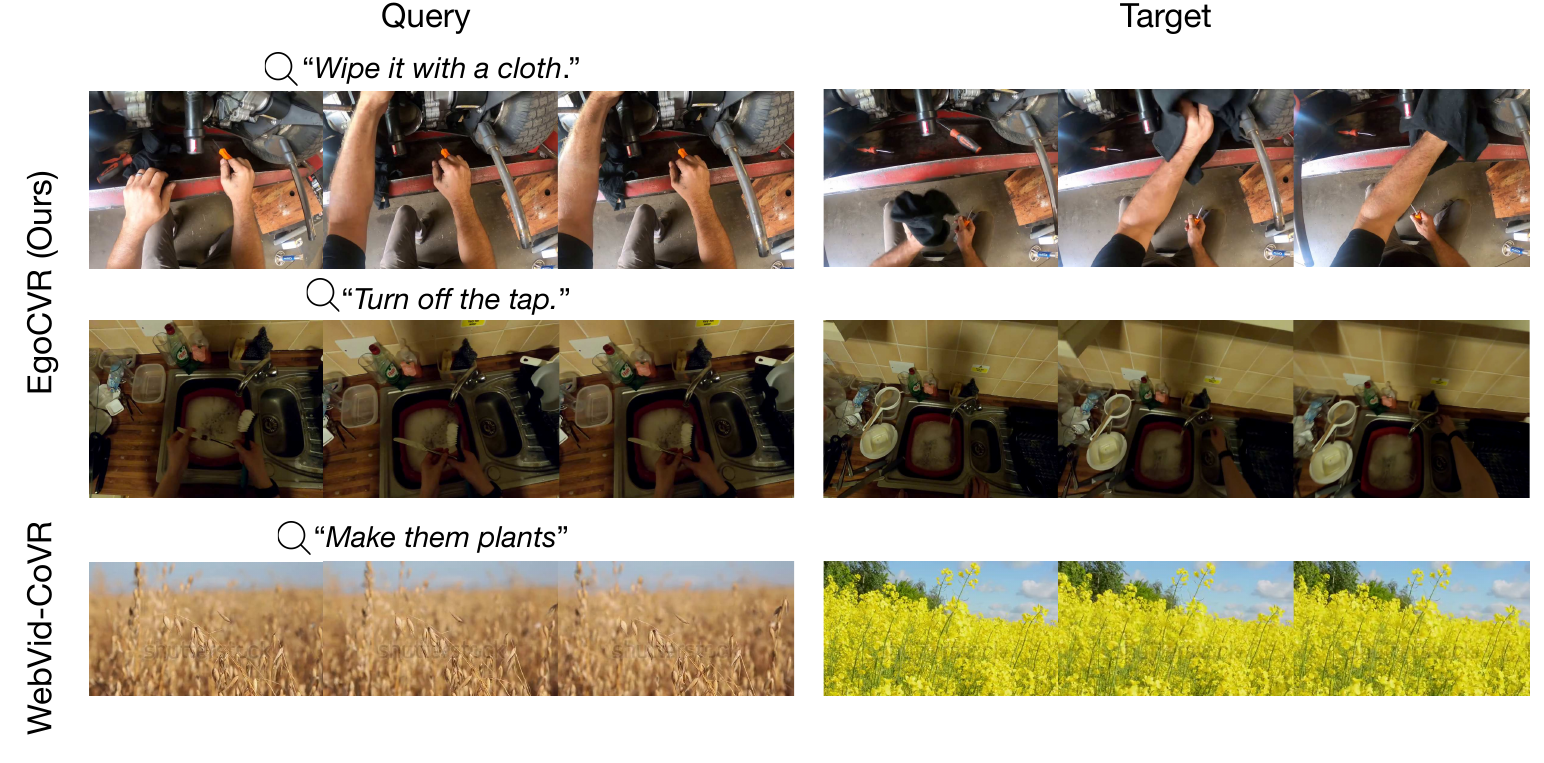}
    \caption{Samples consisting of visual and text queries along with the target video from our test set \benchmarkName (top two rows) and WebVid-CoVR-Test set~\cite{ventura2023covr} (bottom row).}
    \label{fig:dataset_qualitative}
\end{figure}

\myparagraph{Visual Distractors.} We additionally collect distractor video clips for each annotated target video similar to the CIRR image subsets~\cite{cirr}. We automatically source the distractor clips from the same long-form video provided by the Ego4D FHO task. 
Our collected distractor clips ensure high visual similarity (\ie identical camera wearer and scene) and prevent trivial retrieval shortcuts based solely on visual similarity. 
To obtain the distractor clips, for each target video in \benchmarkNameNS, we filter out clips from the Ego4D FHO annotations originating from different long-form videos, clips used as query-target video annotations, and clips depicting the same action as the target. We then rank potential distractor clips by their narration's CLIP similarity to the target video narration. Finally, we sample up to 6 distractor clips per target video.
To represent various semantic similarity levels, we sample one clip from the bottom 10\,\% of similarity scores, four from the middle 80\,\%, and one from the top 10\,\%. %
We sample a total of 10,522 distractors with an average of 4.2 distractors per target video.

\myparagraph{Dataset Statistics.} \benchmarkName is created with the intent to explore the video understanding capabilities of current vision-language models. Our annotation process ensures i) high-quality annotated video pairs and ii) a strong focus on temporal events. 
We analyse the instructions of \benchmarkName and WebVid-CoVR-Test regarding their focus on temporal events. We consider instructions as temporal if the change from query to target video, described using the modification text, directly changes the depicted action or requires temporal video understanding (\ie \texttt{\textit{Pick} it \textit{up} instead.}). In contrast, object-centred modifications require no action understanding but manipulating given objects (\ie \texttt{Cut the \textit{carrot} instead.}). 
To obtain this information, we instruct GPT-4 to assess whether a given instruction focuses on temporal events or objects (see the supplementary for more details). 
Our \benchmarkName evaluation benchmark consists of 2,295 samples, from which 1,811 focus on temporal events (78.9\,\%) and 484 on object-centred changes (21.1\,\%). As visualised in Figure~\ref{fig:dataset_temporal}, this starkly contrasts with WebVid-CoVR-Test, where 85\,\% of samples focus on object-centred modifications.  %

To assess the variety of actions and objects in \benchmarkNameNS, we apply part-of-speech (POS) tagging on the instructions. For actions, we count the occurrences of unique verbs for temporal modifications, while for objects, we count the occurrences of unique direct objects for object-centred modifications. With 179 different actions and 121 unique objects, we find a great variety of actions and objects present in \benchmarkNameNS. %
Video clips in \benchmarkName have a length of 3.9-8.1\,seconds with an average length of 7.9\,seconds. Modification instructions in \benchmarkName are designed with an average of four words to be precise and concise, \ie we ensure that the instruction only describes the transformation and not the target video itself. Instructions vary from short two-word (\eg ``\texttt{Shake it.}'') to longer and more detailed instructions (\eg ``\texttt{Use the other hand to pick up a different object from the shelf.}''). 

\section{Training-Free Re-Ranking Composed Video Retrieval} 
\label{sec:method}

We adopt several vision-language methods for composed video retrieval. 
To show that our proposed benchmark focuses on temporal actions, we employ both image processing and video processing models in the evaluation.
As image processing models, we employ two widely used image-language models, namely \textbf{CLIP}~\cite{clip} and \textbf{BLIP}~\cite{blip}.  
We also adapt the video-language models \textbf{EgoVLPv2}~\cite{egovlpv2} and \textbf{LanguageBind}~\cite{languagebind} that were specifically designed to learn temporal video representations. 
EgoVLPv2 was specifically pre-trained on egocentric videos, while LanguageBind aligns various modalities such as video, infrared, depth and audio to a frozen language encoder after pre-training.
We also employ the recently introduced composed video retrieval framework \textbf{BLIP$_\text{CoVR}$}~\cite{ventura2023covr} and \textbf{BLIP$_\text{CoVR-ECDE}$}~\cite{thawakar2024composed} which leverages the BLIP model for cross-attention between visual and textual encoders. They were specifically finetuned on WebVid-CoVR for CVR, leading to top performance on the WebVid-CoVR test set. We also evaluate CIReVL~\cite{cirevl} on our benchmark since it is a training-free method. %
Below, we only discuss the self-adapted methods  TF-CVR and \methodNameNS.

\myparagraph{Composed Video Retrieval by Language.} We use a methodology very similar to \textbf{CIReVL}~\cite{cirevl}, which has successfully been applied for Composed Image Retrieval. Given a video captioning model such as LaViLa~\cite{lavila}, we can obtain the textual caption of the query video. We name this approach training-free CVR~(\textbf{TF-CVR}). Specifically, given a query video $q_v$, and a video captioning model $\Psi_C$, we obtain its textual representation as $c_q=\Psi_C(q_v)\in\mathcal{T}$. However, this video caption only captures the reference video, not the specified textual modification $q_t$. While the two texts could be combined naively using concatenation, we use an LLM to combine the video caption and textual modifier into a coherent target caption, similar to CIReVL~\cite{cirevl}. Formally, given access to an LLM $\Psi_R$, we generate a target video caption as $c_q^t = \Psi_R(p \circ c_q \circ q_t)$, which queries the LLM with a concatenation of the template prompt $p$, the generated video caption $c_q$ and modification instruction $q_t$. The template prompt $p$ consists of a few in-context examples to guide the LLM and a short task description. Concrete examples of this process are shown in the supplementary material. Given this generated target caption $c_q^t$, TF-CVR searches the video database $\mathcal{D}$ alongside $c_q^t$ using a text-video model (\eg EgoVLPv2~\cite{egovlpv2}, LanguageBind~\cite{languagebind}). The retrieved target $V_q^t$ is:
\begin{equation}
    \label{eq:retrieval}
    V_q^t = \underset{v\in\mathcal{D}}{\mathtt{argmax}}\; \frac{{\Psi_V(v)}^\intercal \Psi_T(c_q^t)}{||\Psi_V(v)|| \cdot ||\Psi_T(c_q^t)||} \quad.
\end{equation}

\myparagraph{Re-Ranking for Composed Video Retrieval.} While the proposed approach, TF-CVR, is simple and effective, a major drawback of the method is that solely relying on text could potentially lead to the selection of semantically similar yet visually unrelated video clips. Therefore, we first apply a visual filtering step to select a candidate video database $D' \subset D$. This is performed by selecting the $n_c$ most similar video clips to the provided reference video $q_v$. This is described more formally as:
\begin{equation}
D' = \underset{v\in\mathcal{D}}{\mathtt{top}n_c} \left( \frac{{\Psi_V(q_v)}^\intercal \Psi_V(v)}{||\Psi_V(q_v)|| \cdot ||\Psi_V(v)||} \right) \quad.
\end{equation}
After filtering, we apply our proposed approach TF-CVR, except now, the video gallery is restricted to $D'$. We refer to this method as training-free re-ranking CVR~(\methodNameNS). We demonstrate the efficacy of this method in Section~\ref{sec:ablation}, especially in settings with a large video gallery. Note that the visual filtering applied in this step can use a different visual encoder than the text-video retrieval step, allowing us to leverage the complementary abilities of different models.

\section{Experiments}

\myparagraph{}%
We explain the two proposed evaluation settings and metrics in Section~\ref{sec:eval}. Further, we discuss the results obtained on \benchmarkName in Section~\ref{sec:res} along with the ablations, analyses and qualitative examples performed on \benchmarkNameNS.

\subsection{Evaluation Settings and Implementation Details}
\label{sec:eval}

\myparagraph{Global and Local Settings.}
We consider two possible evaluation settings for \benchmarkNameNS. The first is the standard composed image/video retrieval setting, where the gallery comprises a long list of videos. We refer to this strategy as the \textit{global} search. In the \textit{global} setting, the query is searched in the pool of videos, which contains all the other video queries, along with their video distractors. Each query tuple has a search gallery of at least 10,661 video clips, with a maximum of 12,526. 
The second setting is the \textit{local} search and is obtained by restricting the gallery to only clips from the same video sequence. This strategy simulates the scenario when searching in a long video for a specific moment. Each query tuple has a gallery of a maximum of 10 clips with an average of 6.4.

\myparagraph{Evaluation Metrics.} We employ the widely used recall metrics, namely Recall@1, Recall@5 and Recall@10 for the \textit{global} setting, while for the \textit{local} setting we employ  Recall@1, Recall@2 and Recall@3, since the length of the gallery is considerably smaller. When a query has more than one target video, we consider the target as true positive only once when one of the target videos is retrieved.

\myparagraph{Implementation Details.}
We perform our experiments using the publicly available official implementations of various vision-language models~\cite{clip,blip,egovlpv2,languagebind}%
, using their default configurations to extract both visual and textual features. %
We employ the ViT-L/14~\cite{vit} version of the CLIP model provided by OpenCLIP~\cite{openclip} which was pre-trained on DataComp-1B~\cite{gadre2024datacomp}, as well as, the BLIP-Large variant finetuned on COCO~\cite{coco} and the BLIP model finetuned on WebVid-CoVR~\cite{ventura2023covr} by Ventura \etal~\cite{ventura2023covr}. We use the fully finetuned video encoder for LanguageBind~\cite{languagebind} and EgoVLPv2~\cite{egovlpv2} with full projection.
Unless otherwise noted, we use $n_c=15$ and employ EgoVLPv2 as the text encoder $\Psi_T$ in \methodName (Equation~\ref{eq:retrieval}).
CLIP and BLIP visual representations for the videos are obtained by averaging embeddings from 15 uniformly sampled image frames.

\begin{table*}[t]
\centering
\setlength\tabcolsep{4.0pt}
\caption{Results on both the global and local evaluation settings on \benchmarkNameNS. Our proposed \methodName achieves state-of-the-art results in both the global and local settings. We also report several baselines that only use the text, the reference video, or a naive average of the visual and textual embeddings (Fusion Strategy Avg). The best and the second best results are in \textbf{bold} and \underline{underlined}, respectively.}%
\resizebox*{1.0\textwidth}{!}{%

\begin{tabular}{lcccc|ccc|ccc}
\toprule

\multirow{2}{*}{Method}  & Video  & Textual & Visual & Fusion &\multicolumn{3}{c}{Global} & \multicolumn{3}{c}{Local}   \\
  & Model & Input & Input & Strategy & R@1 & R@5 & R@10 & R@1 & R@2 & R@3 \\

\midrule
Random & \xmark & \xmark & \xmark & -  & 0.01 &  0.05 &  0.1 & 25.3 & 38.2 & 50.7 \\ \midrule
CLIP & \xmark & \cmark & \xmark & -   & 0.7 & 1.7 & 2.7 & 33.5 & 48.8 & 61.8 \\
BLIP & \xmark & \cmark & \xmark & -   & 0.4 & 1.4 & 2.7 & 32.5 & 46.9 & 59.7 \\
EgoVLPv2 & \cmark & \cmark & \xmark & -   & 1.7 & 3.9 & 7.2 & \underline{41.0} & \underline{57.3} & \underline{69.0} \\
LanguageBind & \cmark & \cmark & \xmark & -   & 0.9 & 2.7 & 4.2 & 34.2 & 51.1 & 64.1 \\

\midrule
CLIP & \xmark & \xmark & \cmark & -  & 7.4 & 33.2 & \underline{55.3} & 26.1 & 43.4 & 57.7 \\
BLIP & \xmark & \xmark & \cmark & -  & 6.5 & 32.6 & \underline{55.3} & 26.5 & 43.7 & 57.5 \\
EgoVLPv2 & \cmark & \xmark & \cmark & -  & 7.6 & 32.5 & 49.6 & 27.5 & 44.3 & 59.1 \\
LanguageBind & \cmark & \xmark & \cmark & -  & 6.1 & 33.1 & 53.4 & 26.1 & 42.9 & 57.7 \\

\midrule
CLIP & \xmark & \cmark & \cmark & Avg  & 7.5 & 33.6 & \textbf{55.6} & 26.4 & 43.7 & 57.9 \\
BLIP & \xmark & \cmark & \cmark & Avg  & 8.7 & 32.9 & 52.8 & 29.5 & 45.9 & 61.0 \\
EgoVLPv2 & \cmark & \cmark & \cmark & Avg  & \underline{9.5} & \underline{34.9} & 52.1 & 30.7 & 51.3 & 66.0 \\
LanguageBind & \cmark & \cmark & \cmark & Avg  & 6.1 & 33.2 & 53.5 & 26.1 & 43.1 & 57.8 \\
 
\midrule
BLIP$_\text{CoVR}$~\cite{ventura2023covr} & \xmark & \cmark & \cmark & Cross-Attention  & 5.4 & 15.2 & 24.3 & 33.1 & 49.5 & 62.9 \\
BLIP$_\text{CoVR-ECDE}$~\cite{thawakar2024composed} & \xmark & \cmark & \cmark & Cross-Attention  & 6.0 & 16.3 & 24.8 & 33.4 & 49.3 & 63.0 \\
CIReVL~\cite{cirevl} & \xmark & \cmark & \cmark & Captioning  & 2.0 & 6.8 & 10.6 & 33.6 & 49.7 & 61.4 \\
\methodName (Ours) & \cmark & \cmark & \cmark & Captioning  & \textbf{14.1} & \textbf{39.5} & {54.4} & \textbf{44.2} & \textbf{61.0} & \textbf{73.2} \\

\bottomrule
\end{tabular}
}

\label{tab:allmain}
\end{table*}
 
\subsection{Benchmark Evaluation and Model Ablations} 
\label{sec:res}

We explore the potential of different query modalities in fine-grained composed video retrieval for \benchmarkNameNS. Specifically, we use three methods for video ranking: retrieval using only text query (\textbf{text-only}), retrieval using only visual query (\textbf{visual-only}), and retrieval using both text and visual queries (\textbf{visual-text}). %

\myparagraph{Global Setting.} The results for the global evaluation setting are presented in Table \ref{tab:allmain}. We notice the absolute performance of the Recall@k (k $\in \{1, 5, 10\}$) values being quite low due to having thousands of candidate video clips in the gallery for each query. However, we observe that relying solely on the text performs extremely poorly for all methods, as they fail to achieve an R@1 of even 2\%. We notice that methods relying on visual features demonstrate competitive performance (up to 7.6\% in R@1). In this setting, our proposed \methodName achieves the best results (R@1 of 14.1\%) due to the combination of LanguageBind-based candidate filtering using visual features, followed by re-ranking using the generated target caption. It is also notable that the BLIP finetuning methods (BLIP$_\text{CoVR}$ and BLIP$_\text{CoVR-ECDE}$, which attain state-of-the-art results on WebVid-CoVR as well as several CIR benchmarks, does not generalise to our proposed \benchmarkName benchmark, achieving an R@1 value of only 5.4\% and 6.0\% respectively. We also observe that CIReVL~\cite{cirevl}, which was tailored for the task of CIR, does not generalise directly to videos, obtaining an R@1 score of 2\%.

\myparagraph{Local Setting.} In the local setting, we notice diminishing returns from methods that rely solely on visual features, performing only marginally better than random selection. This is due to the fact that, in this setting, all the videos in the gallery are by design very similar. Therefore, the visual similarity is not too helpful. Textual search performs much better because in the local setting, the textual information is the main indicator in finding the videos in the gallery. %
\methodName performs text-video retrieval with a full caption that also captures the information from the source video, achieves the  best result (R@1 of 44.2\%).

\begin{table}[t]
\centering
\begin{minipage}{0.44\textwidth}
\caption{Results in terms of R@1, R@5 and R@10 on the global setting that emphasize the importance of temporal information on \benchmarkNameNS.}
\vspace{-5pt}
\centering
\setlength\tabcolsep{4.0pt}
\resizebox{\textwidth}{!}{%
\begin{tabular}{lc|ccc}
\toprule
{Method} & Temporal & R@1 & R@5 & R@10 \\
\midrule
\multirow{2}{*}{LanguageBind} & \xmark & \textbf{6.4} & 25.3 & 38.0  \\
   &   \cmark & 6.1 & \textbf{33.1} & \textbf{53.4}  \\
\midrule
\multirow{2}{*}{EgoVLPv2} &  \xmark   & 6.9 & 24.0 & 34.8  \\
  &    \cmark          & \textbf{9.5} & \textbf{34.9} & \textbf{52.1}  \\
\midrule
\multirow{2}{*}{BLIP$_{\text{CoVR}}$} &  \xmark  & 4.1 & 12.5 & 19.6  \\
 & \cmark & \textbf{5.4} & \textbf{15.2} & \textbf{24.3}  \\
\midrule
\multirow{2}{*}{\methodNameNS} &  \xmark & 10.2 & 27.4 & 39.2  \\
 &   \cmark  & \textbf{14.1} & \textbf{39.5} & \textbf{54.4}  \\
\bottomrule
\end{tabular}
}

\label{tab:temporal}
\end{minipage}
\hfill
\begin{minipage}{0.54\textwidth}
\caption{Results in terms of R@1, R@5 and R@10 demonstrating the importance of the two-stage (filtering and re-ranking) process for our proposed \methodName on the global setting of \benchmarkNameNS. When applying re-ranking, we show the model that is applied for visual filtering before using TF-CVR.}
\centering
\setlength\tabcolsep{4.0pt}
\resizebox{\textwidth}{!}{%
\begin{tabular}{lccc}
\toprule
{Method}  & R@1 & R@5 & R@10 \\
\midrule
LanguageBind (Stage 1) & 6.1 & 33.1 & 53.4 \\
EgoVLPv2 (Stage 1) & 7.6 & 32.5 & 49.6 \\
TF-CVR (Stage 2) & 4.4 & 12.9 & 18.3 \\
\methodName (EgoVLPv2 $\rightarrow$ TF-CVR) & 12.2 & 35.1 & 49.5 \\
\methodName (LanguageBind $\rightarrow$ TF-CVR) & \textbf{14.1} & \textbf{39.5} & \textbf{54.4} \\
\bottomrule
\end{tabular}

}

\label{tab:reranking}
\end{minipage}
\end{table}

\myparagraph{Benefits of Temporal Information.} We demonstrate the benefits of using temporal information through processing the whole video compared to using only a single frame sampled from the middle of the video. The results are reported in Table~\ref{tab:temporal}. %
We observe that temporal information improves performance significantly across all methods (up to $12.1$ percentage points in terms of R@5), confirming that our benchmark benefits and requires temporal understanding.
This is particularly noticeable on the R@5 and R@10 metrics, where we observe \methodName improving from 27.4\% to 39.5\% and from 39.2\% to 54.3\%, respectively, emphasising the importance of using temporal information for this task. 

\begin{figure}[t]
    \centering
    \includegraphics[width=0.99\textwidth]{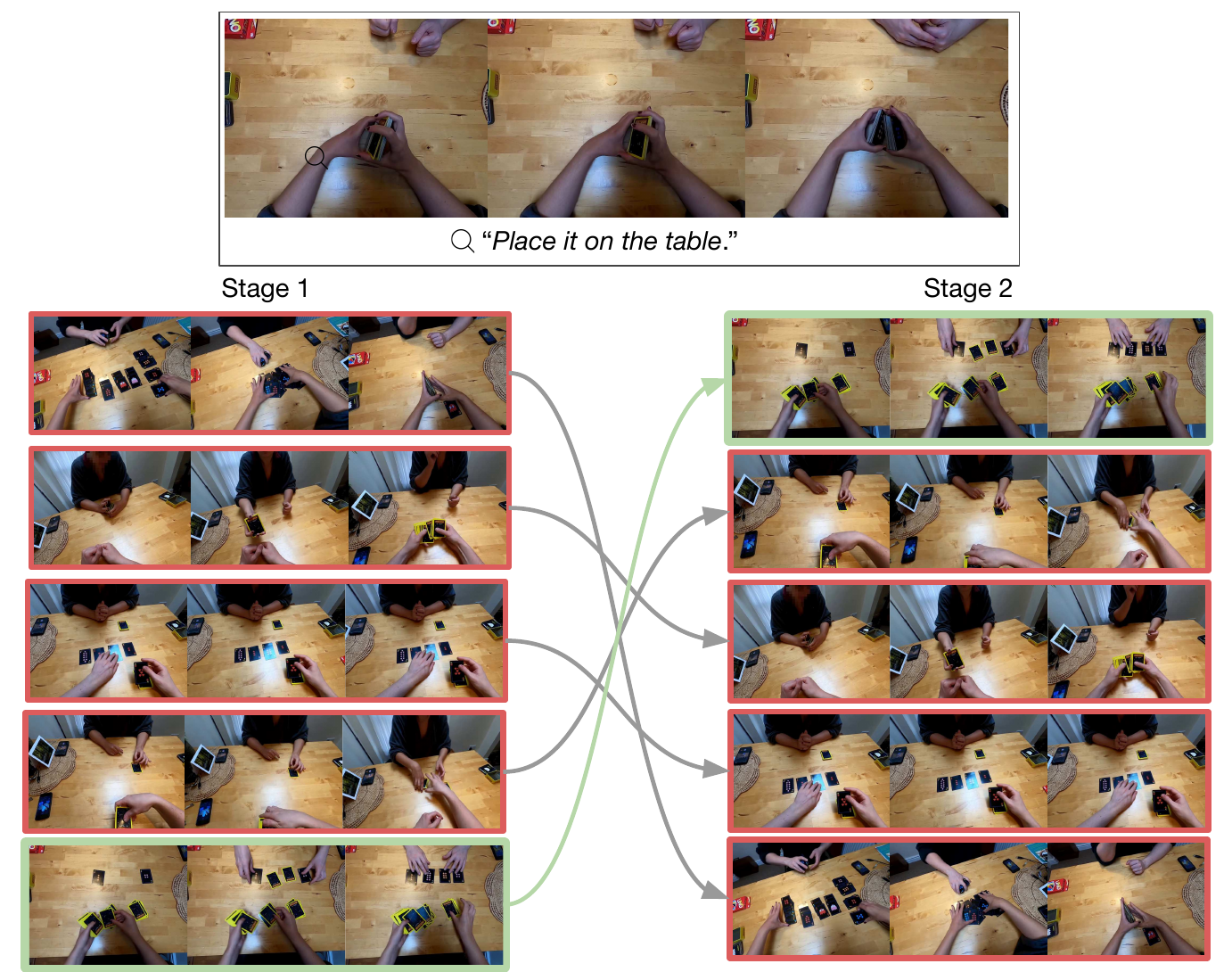}
    \caption{The first and the second stage ranking results of the \methodName method. The arrows indicate how the ranking was changed. The correct video is showcased in green.} %
    \label{fig:1vs2stage}
\end{figure}

\begin{wrapfigure}{r}{0.45\textwidth} 
    \centering
    \includegraphics[width=0.45\textwidth]{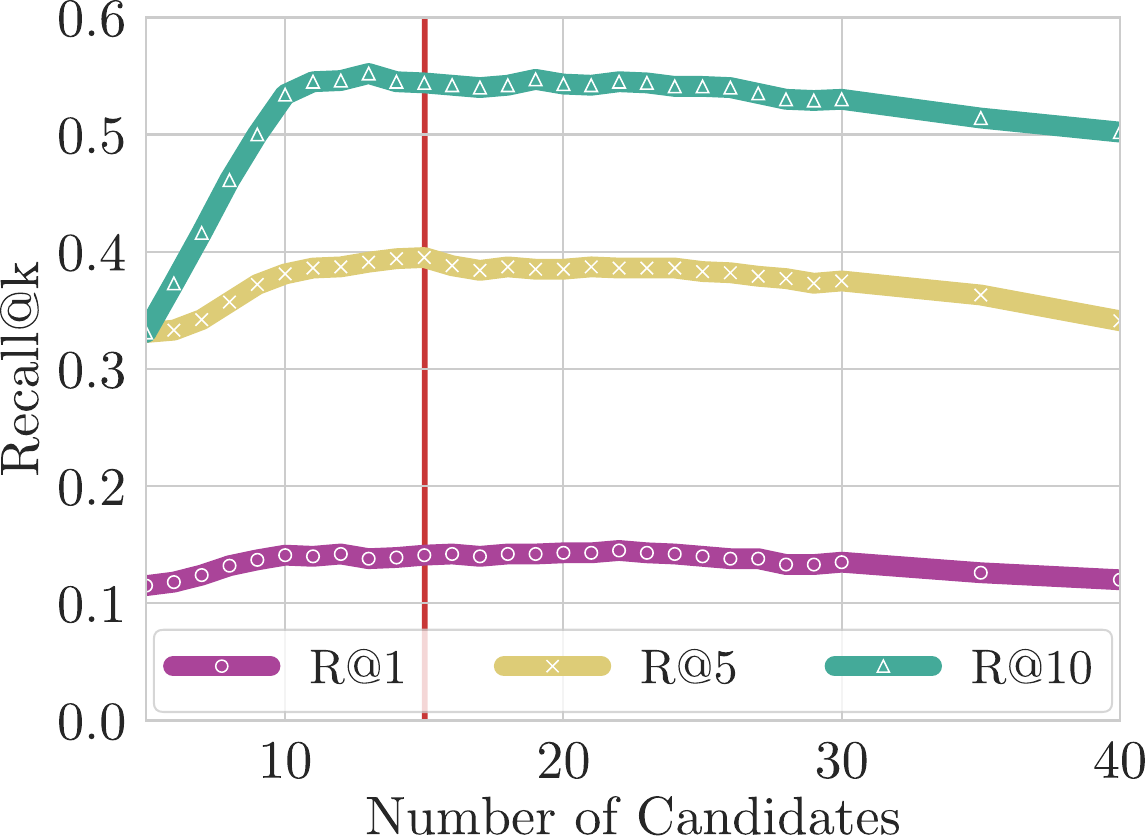}
    \caption{
    Effect of the number of candidates $n_c$ for the visual re-ranking step of \methodNameNS. The vertical line denotes the value of $n_c$ used in our experiments.}
    \label{fig:neighbors}
\vspace{-5pt}
\end{wrapfigure}
\label{sec:ablation}
\myparagraph{Benefits of Re-Ranking.} We also demonstrate the efficacy of our two-stage approach, \methodName (\ie first selecting candidates using visual similarity and re-ranking them using text similarity), on the global setting in Table~\ref{tab:reranking}. 
We notice that only using visual similarity or textual search alone is insufficient, while combining the two steps leads to the best-performing results (last two rows). Additionally, we highlight the benefits of \methodName in drawing complementary knowledge from distinct models. For instance, TF-CVR employs the textual encoder from EgoVLPv2. Using LanguageBind for visual filtering in \methodName (last row) instead of EgoVLPv2 improves the retrieval results across all metrics.

\noindent In Figure~\ref{fig:1vs2stage}, we illustrate the resulting order of the videos obtained after re-ranking.
We can notice that in the first stage, while all videos are visually similar, the correct video is ranked lower, while after the second stage, the target video is moved to the first position, resulting in a correct retrieval.

\begin{table*}[ht]
\centering
\setlength\tabcolsep{4.0pt}
\caption{Text-only retrieval results obtained with CLIP~\cite{clip}, LanguageBind~\cite{languagebind}, and \methodName on \benchmarkNameNS. 
As text query alternatives, we switch among the \textit{instruction}, the caption prediction from video captioning~\cite{lavila} combined with LLM reformulation (\textit{Pred.\ Caption}), and lastly the ground-truth narration (\textit{GT Caption}) available from Ego4D. The video query is used by \methodName for visual re-ranking on the global evaluation.}
\begin{tabular}{lr|ccc|ccc}
\toprule

\multirow{2}{*}{Method} & \multirow{2}{*}{Text Source} & \multicolumn{3}{c}{Global} & \multicolumn{3}{c}{Local} \\   
  & & R@1 & R@5 & R@10 & R@1 & R@2 & R@3 \\

\midrule
\multirow{3}{*}{CLIP} & Instruction  &  0.7 & 1.7 & 2.7 & 33.5 & 48.8 & 61.8 \\
 & Pred.\ Caption  &  1.5 & 4.2 & 7.5 & 34.0 & 49.8 & 63.7 \\
 & GT Caption  &  2.1 & 5.9 & 9.3 & 35.1 & 52.0 & 69.4 \\
\midrule
\multirow{3}{*}{LanguageBind} & Instruction & 0.9 & 2.7 & 4.2 & 34.2 & 51.1 & 64.1 \\
 & Pred.\ Caption & 1.7 & 5.7 & 8.2 & 36.6 & 52.2 & 64.8 \\
 & GT Caption & 3.3 & 8.0 & 11.5 & 39.2 & 56.6 & 69.1 \\
\midrule
\multirow{3}{*}{\methodNameNS} & Instruction & 12.8 & 35.3 & 53.4 & 41.0 & 57.3 & 69.0  \\ 
 & Pred.\ Caption & 14.1 & 39.5 & 54.4 & 44.2 & 61.0 & 73.2  \\ 
 & GT Caption & 18.5 & 44.7 & 58.5 & 51.7 & 69.7 & 81.8  \\ 
  
\bottomrule
\end{tabular}

\label{tab:gt}
\end{table*}

\myparagraph{Effect of the Number of Re-Ranking Candidates.} 
Our approach involves re-ranking the candidates chosen by the first stage of visual filtering. We study the impact of the number of neighbours chosen after the first stage in Figure~\ref{fig:neighbors}.
We observe that the performance stops fluctuating once we select a sufficient number of candidates $n_c$ for re-ranking ($n_c > 10$).
Once the number of candidates becomes too large ($n_c > 30$), the performance starts diminishing and eventually loses the benefits of the visual filtering. 
In our experiments, we use $n_c=15$. However, the final results are stable within a large range of selected candidates. 

\myparagraph{Effect of Text-Caption.} We also investigate the benefits of using an LLM to generate a plausible caption of the video, along with its shortcomings in Table~\ref{tab:gt}. This is achieved by resorting to text-video retrieval with different textual inputs. We experiment with the input textual modification, the predicted caption (as the result of video captioning and LLM reformulation described in Section~\ref{sec:method}), as well as using the ground-truth target caption provided by Ego4D. The ground-truth target caption serves as a useful upper bound on text-only search. We notice that the video captioning and LLM reformulation consistently improve the results for all the models. In the global setting, the improvement is especially noticeable as the R@1 result increases at least twice (from $0.7\%$ to $2.1\%$ for the CLIP model) due to having a complete caption instead of a brief text. While the ground truth caption naturally improves the results further, it must be noted that improving the caption further does not offer much room for improvement. For instance, in the local setting, the R@1 result improves from 44.2\% to 51.7\% when our method is employed. Future work on \benchmarkName would benefit both from improving the underlying text-video models through better foundation models~\cite{zhao2024videoprism,videollava}, as well as, from developing methods that can cohesively utilise the reference video and the textual instruction simultaneously.  

\begin{figure}[t!]
    \centering
    \includegraphics[width=0.99\textwidth]{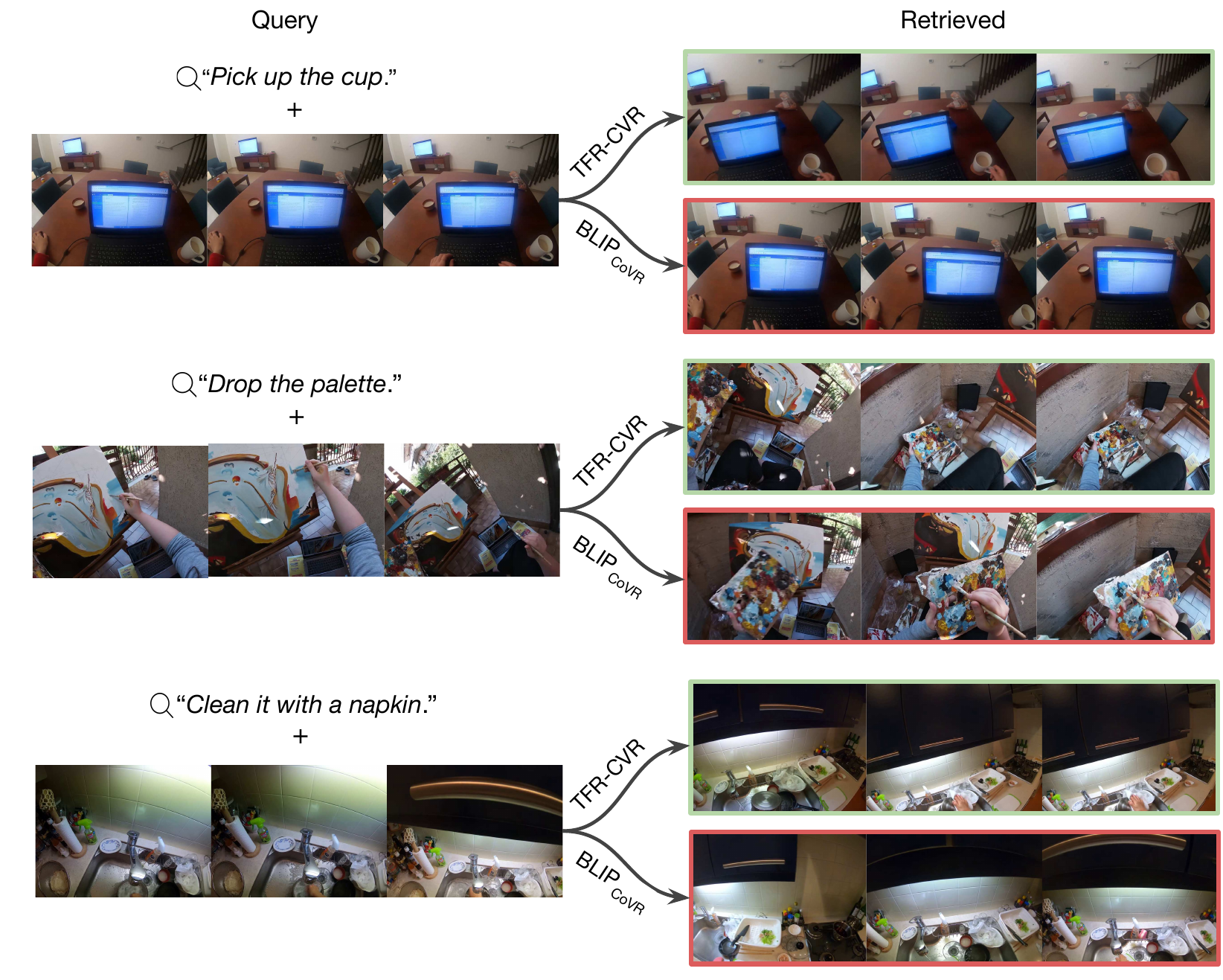}
    \caption{Qualitative examples of composed video retrieval ranking on \benchmarkNameNS. For each example, we show the queries along with the clip retrieved by \methodName and BLIP$_\text{CoVR}$~\cite{ventura2023covr}. The target videos are enclosed in green rectangles.
    }
    \label{fig:qualitative_examples}
\end{figure}

\myparagraph{Qualitative Examples.} We demonstrate the benefits of our re-ranking approach in Figure~\ref{fig:1vs2stage}. We observe that relying on visual features to select the top candidates results in visually similar clips, without capturing subtle actions accurately. After re-ranking with the text-video retrieval using the predicted caption, fine-grained actions are accurately captured in the final ranking.  %

\noindent We additionally illustrate qualitative examples in Figure~\ref{fig:qualitative_examples}, comparing the retrieved samples of the \methodName and BLIP$_{\text{CoVR}}$ methods. We observe that our proposed method performs better than BLIP$_{\text{CoVR}}$, retrieving all targets. Notably, these examples require fine-grained action understanding. While \methodName returns the correct clip, BLIP$_{\text{CoVR}}$ retrieves visually similar clips, however, they do not display the correct action. This highlights the inherent limitations of an image-based model despite being finetuned for Composed Video Retrieval.

\myparagraph{Limitations.} We provide a high-quality evaluation benchmark for CVR. However, collecting a training set, even through an automated process, would allow finetuning models for CVR instead of adapting existing vision-language models in a training-free manner. Furthermore, our evaluation benchmark also consists of egocentric videos, however, it can be employed to assess the generalization of any CVR model. Despite the aforementioned limitations, we believe that our proposed benchmark serves as an intriguing validation ground for adapting existing vision-language models and a valuable evaluation set for high-quality temporal action understanding. Expanding the scope of the benchmark to include different types would also increase the diversity and applicability of the findings. %

\section{Conclusion}
In this work, we introduce the \benchmarkName benchmark for the task of fine-grained Composed Video Retrieval. We demonstrate that existing text-video and Composed Video Retrieval methods do not directly generalise to \benchmarkNameNS. Therefore, we introduce our method \methodNameNS, which uses existing video and language models in a modular fashion to achieve strong results on \benchmarkNameNS. We also show the shortcomings of existing vision-language models, even when they are explicitly finetuned for Composed Video Retrieval. We hope that our benchmark and method inspire further work on fine-grained action understanding and retrieval. 
\subsection*{Acknowledgements}
This work was supported by BMBF FKZ: 01IS18039A, by the ERC (853489 - DEXIM), by EXC number 2064/1 – project number 390727645.
Thomas Hummel and Shyamgopal Karthik thank the International Max Planck Research School for Intelligent Systems (IMPRS-IS) for support. We also thank Yavuz Durmazkeser for his assistance with data labelling, which contributed to increasing our data size. 

\bibliographystyle{splncs04}
\bibliography{egbib}

\appendix

\title{Supplementary Material} 

\titlerunning{Supplementary Material EgoCVR}
\author{}
\authorrunning{T. Hummel et al.}
\institute{}
\maketitle

In this appendix, we report results when applying re-ranking to other methods in Section~\ref{supp:re-ranking-other}, show failure cases and future directions of \methodName in Section~\ref{supp:failure}, and report results with \methodName on WebVid-CoVR-Test~\cite{ventura2023covr} in Section~\ref{supp:results-webvid}.
Furthermore, we provide more details on \benchmarkNameNS's diversity (Section~\ref{diversity}), present the instruction prompts given to the LLM models to create the text modification (Section~\ref{instruction}), to create the target captions for \texttt{TF-CVR} (Section~\ref{target}) and to perform the temporal event detection analysis on the CVR benchmarks (Section~\ref{temporal}), as well as provide additional qualitative examples (Section~\ref{illustrations}). %

\section{Re-Ranking Applied to Other Methods}\label{supp:re-ranking-other}
We show the results for applying re-ranking using the same LanguageBind encoder for all the methods below. While the re-ranking improves BLIP$_\text{CoVR}$, the overall performance and improvement are much larger for \methodNameNS.%
\begin{table}[h]
\vspace{-0.5em}
\centering
\setlength\tabcolsep{4.0pt}
\caption{Results on \benchmarkName in terms of R@1, R@5 and R@10 on the global setting with and without applying re-ranking. We also report the mean recall change when applying the re-ranking. }%
\small
\begin{tabular}{l|ccc|ccc|c}
\toprule
\multirow{2}{*}{Method} &\multicolumn{3}{c}{w/o re-ranking} & \multicolumn{3}{c|}{w/ re-ranking} & \\
   & R@1 & R@5 & R@10 & R@1 & R@5 & R@10 & $\Delta$ R@\{1,5,10\}\\
\midrule
CLIP~\cite{clip} & 7.5 & 33.6 & \textbf{55.6} & 7.5 & 33.5 & 54.4 & \textcolor{myRed}{$0.4\downarrow$}\\
BLIP~\cite{blip} & 8.7 & 32.9 & 52.8 & 8.7 & 33.8 & 54.2 & \textcolor{myGreen}{$0.8\uparrow$}\\
BLIP$_\text{CoVR}$~\cite{ventura2023covr} & 5.4 & 15.2 & 24.3 & 10.4 & 31.9 & 52.2 & \textcolor{myGreen}{$16.5\uparrow$}\\
\midrule
\methodName & {4.4} & {12.9} & {18.3} & {14.1} & {39.5} & {54.4} & \textcolor{myGreen}{$\textbf{24.1}\uparrow$}\\
\bottomrule
\end{tabular}
\vspace{-1em}
\label{sup:reranking}
\end{table}

\section{Failure Cases and Future Directions}\label{supp:failure}
\begin{figure}[h]
\centering
\includegraphics[width=\columnwidth]{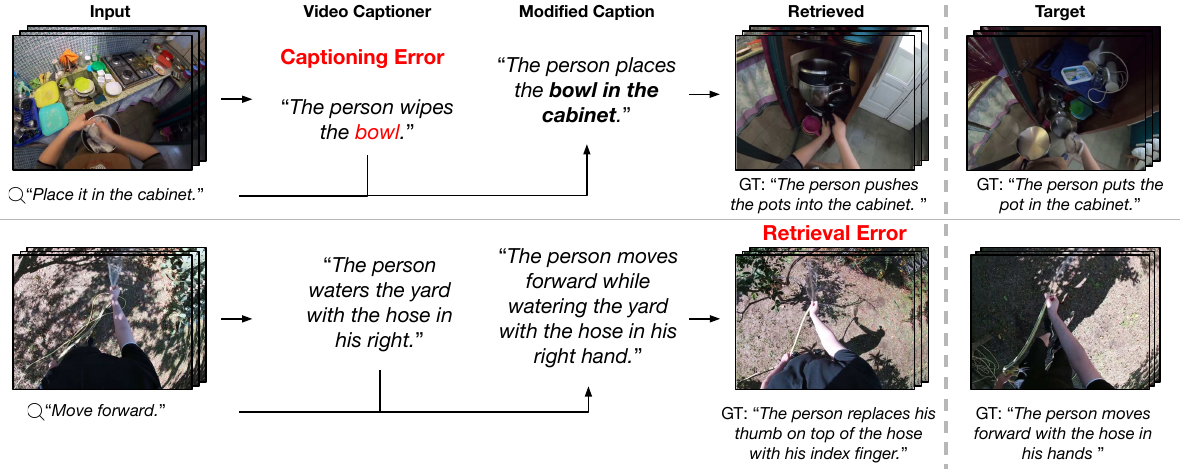}
\caption{Qualitative depiction of failure cases of \methodNameNS. The modular approach of \methodName allows us to trace back failure cases mostly to two main sources: video \textit{captioning errors} (top) and text-to-video \textit{retrieval errors} (bottom).}
\label{sup:failure}
\end{figure}

Due to the modular approach of \methodNameNS, we can break down the failure cases. As shown in Table\,4 in the main paper, employing the ground-truth (GT) captions achieves only an R@1 of 51.7\% with a gallery of 7 candidates (Local).
Therefore, the biggest source of improvement on \benchmarkNameNS\ would be from applying stronger text-to-video retrieval models. We can also trace back errors to erroneous video captions.
In Figure~\ref{sup:failure}, we show an example of the wrong retrieval caused by the text-to-video retrieval method at the bottom. In addition, we highlight an error caused by the video captioning method on the top. Here, the \textit{``pot''} was mistaken for a \textit{``bowl''} during the captioning, leading to the retrieval of a video that mainly depicts a bowl.

\section{Results on WebVid-CoVR Benchmark}\label{supp:results-webvid}
We additionally show results of \methodName on the WebVid-CoVR-Test~\cite{ventura2023covr} benchmark in Table~\ref{sup:webvid}. For \methodNameNS, we employ LanuguageBind~\cite{languagebind} as the visual and textual encoder, since it was trained on different types of videos, unlike EgoVLPv2~\cite{egovlpv2} which is restricted to egocentric videos. We observe that \methodName is able to achieve the best results among all methods that do not explicitly train on the WebVid-CoVR training set. The performance (R@1 of 51.7\%) is also quite competitive with the state-of-the-art result obtained by BLIP$_{\text{CoVR}}$~\cite{ventura2023covr} (R@1 of 53.1\%). We also continue to notice a constant benefit from applying the re-ranking strategy. After applying the visual-only filtering of LanguageBind (which in itself is a weak model scoring an R@1 value of 43.2\%), the performance of the text-based \texttt{TF-CVR} method improves from R@1 of 48.4\% to 51.7\%. 
\begin{table*}[h]
\centering
\setlength\tabcolsep{4.0pt}
\caption{Results on WebVid-CoVR-Test~\cite{ventura2023covr} in terms of Recall@1, Recall@5 and Recall@5. We report results with models evaluated in the Zero-Shot setting~\cite{clip, blip, languagebind} (including our TF-CVR and \methodNameNS) and with models specifically trained for this data set~\cite{ventura2023covr}.}%
\resizebox*{1.0\textwidth}{!}{%

\begin{tabular}{lcc|ccc}
\toprule

\multirow{2}{*}{Method}  & Zero- & Fusion & \multirow{2}{*}{R@1} & \multirow{2}{*}{R@5} & \multirow{2}{*}{R@10}\\
& Shot & Strategy &  & &  \\

\midrule
Random & \cmark & -  & 0.08 &  0.23 &  0.35 \\ \midrule

CLIP~\cite{clip} & \cmark & Avg & 44.4 & 69.1 & 77.7 \\
BLIP~\cite{blip} & \cmark & Avg & 45.5 & 70.5 & 79.5 \\ 
LanguageBind~\cite{languagebind} (stage 1; visual) & \cmark & - & 43.2 & 66.3 & 75.2 \\
TF-CVR & \cmark & Captioning & 48.4 & 73.7 & \textbf{81.9} \\
\methodName (LanguageBind $\rightarrow$ TF-CVR) ($n_c=20$) & \cmark & Captioning & \textbf{51.7} & \textbf{75.3} & 80.7 \\ 
\midrule

\color{myGray}{BLIP$_\text{CoVR}$~\cite{ventura2023covr}} & \xmarkg & \color{myGray}{Cross-Attention} & \color{myGray}{53.1} & \color{myGray}{79.9} & \color{myGray}{86.9} \\
\bottomrule
\end{tabular}
}
\label{sup:webvid}
\end{table*}

\section{Additional Details to \benchmarkNameNS}

\subsection{Dataset Diversity}\label{diversity}
\begin{figure}[h]
\vspace{-0.5em}
    \begin{subfigure}[h]{0.60\columnwidth}
    \includegraphics[width=\columnwidth,trim={0 8.5cm 0 0},clip]{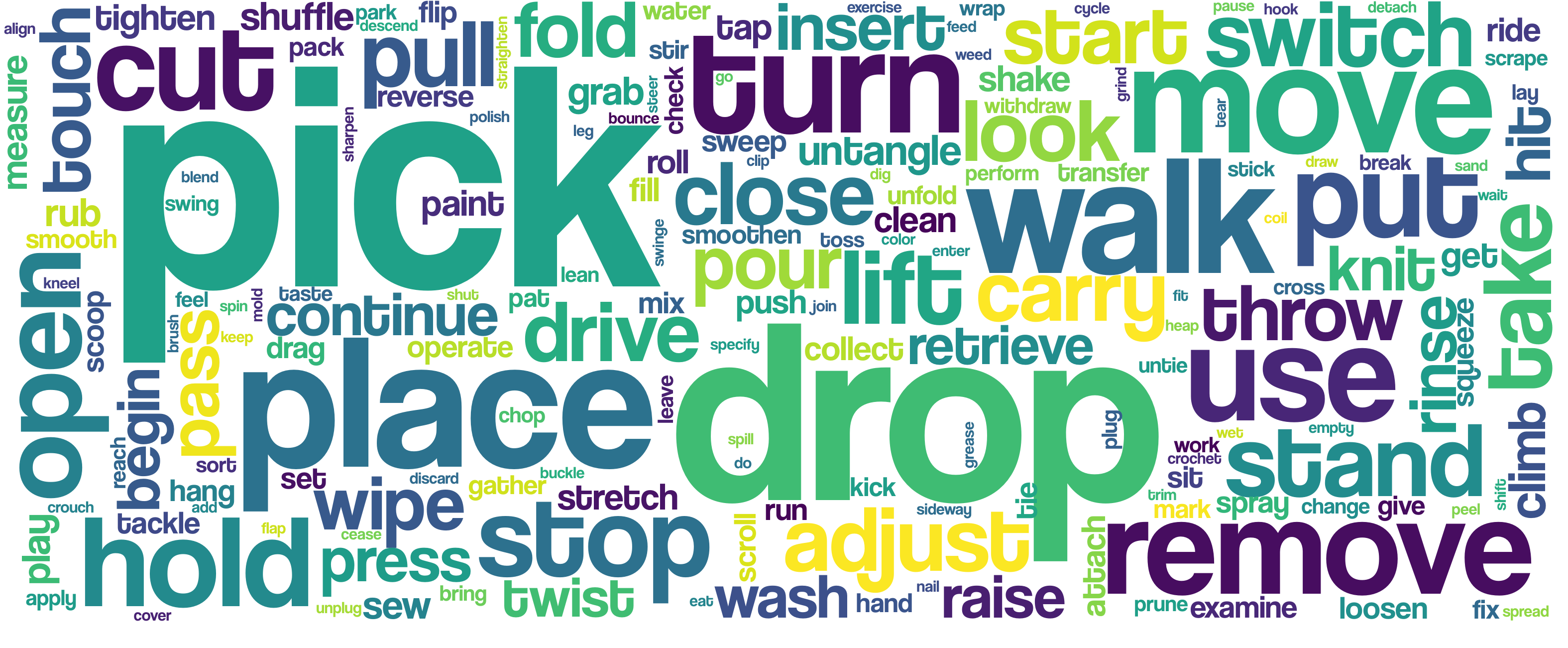}
    \caption{Diversity in actions}
    \label{fig:wordcloud_action}
    \end{subfigure}
    \rulesep
    \begin{subfigure}[h]{0.37\columnwidth}
\includegraphics[width=\columnwidth]{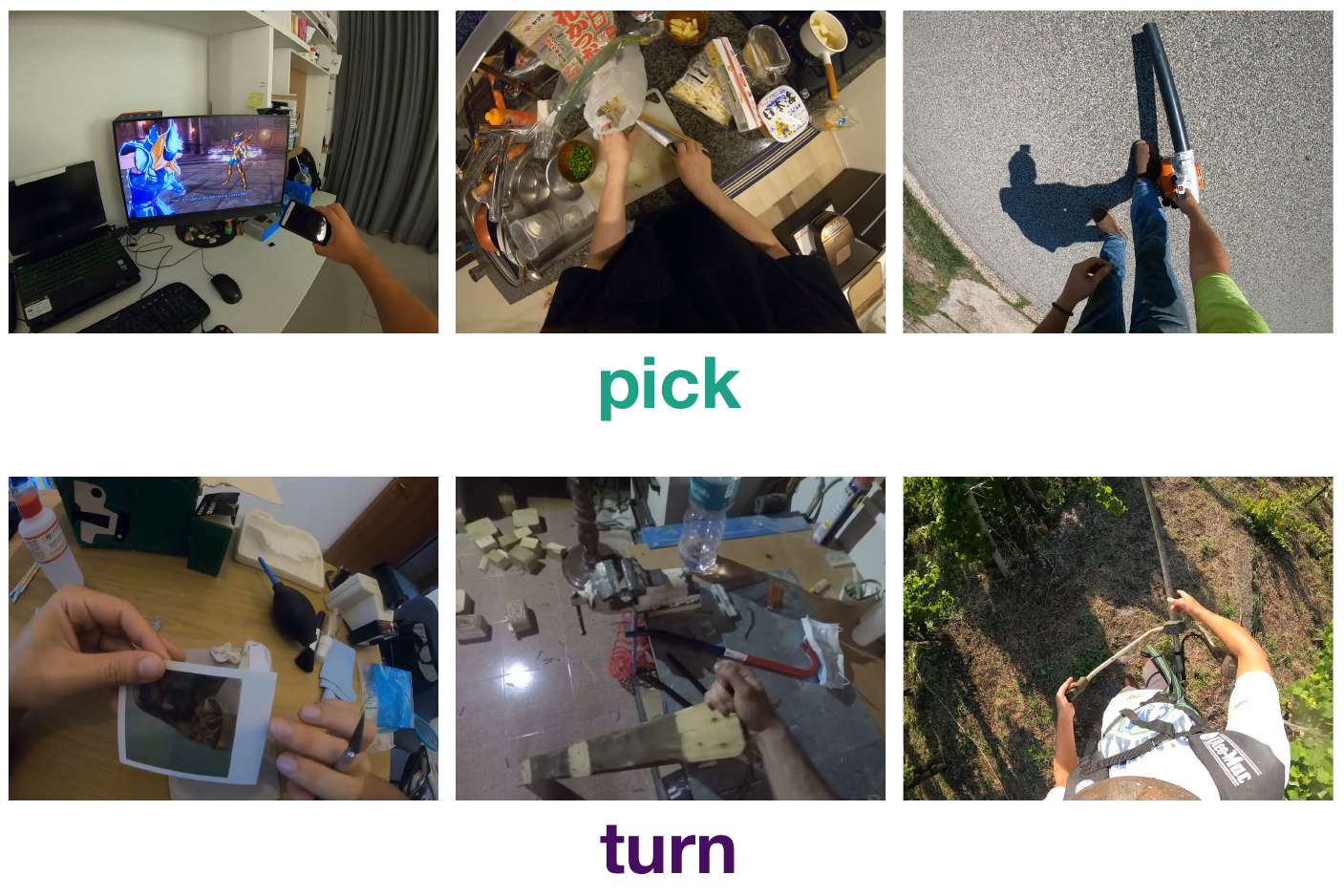}
    \caption{Diverse environments}
    \label{fig:wordcloud_action}
    \end{subfigure}
\caption{Diversity of actions and environments in \benchmarkNameNS.}
\label{fig:diversity}
\vspace{-1em}
\end{figure}

\benchmarkNameNS\ entails 179 different actions across 47 distinct scenarios (top-6: \textit{crafting, cooking, cleaning, construction, carpenter, car mechanic}). 
In Figure~\ref{fig:diversity}, we show (i) a word cloud illustrating the various actions included in \benchmarkNameNS, and (ii) exemplary of diversity in action environments (``\textit{pick}'' appears in 30 and ``\textit{turn}'' in 9 scenarios). \benchmarkName offers wide-ranging environments, objects and actions.

\subsection{Creating Video Modification Instructions}\label{instruction} 
In this section, we give more details on how we create video modification instructions. We automatically generate instructions from query and target clip narrations. More specifically, given the narrations of the source and the target video, the goal is to generate the instruction (modification) text. To achieve this, we use 15 in-context examples showing how this could be done effectively, focusing only on the modification and not the source and target captions itself. 

\begin{tcolorbox}[
enhanced,
colback=white!98!gray,
colframe=black, 
boxrule=.5pt,
colbacktitle=white!88!gray,
coltitle=black, 
fonttitle=\bfseries,
title={Dataset Generation (Instruction) Prompt}, 
breakable=true,
]
\scriptsize
I have 2 videos. Given a brief description of the source and the target video, write an instruction that describes the transformation from the source to the target. The caption you generate should only talk about the necessary modifications. Keep the instruction as short as possible, and focus always on the action. Mention objects only when absolutely necessary. You should not describe objects common to both descriptions, instead use pronouns. Describe only the transformation required. Use the examples below for reference.\\\\
\textbf{Source Narration: }\#C C picks up the jug.\\
\textbf{Target Narration: }\#C C cleans the jug.\\
\textbf{Instruction: }The person is cleaning.\\\\
\textbf{Source Narration: }\#C C picks a spanner from the table with his right hand.\\
\textbf{Target Narration: }\#C C picks a gasket from a table with his right hand.\\
\textbf{Instruction: }Gasket being picked up.\\\\
\textbf{Source Narration: }\#C C picks up the wood from the shelf with his left hand\\
\textbf{Target Narration: }\#C C detaches a wood from the wooden structure with his right hand\\
\textbf{Instruction: }Person uses the other hand and detaches.\\\\
\textbf{Source Narration: }\#C C fixes the bolt on the motorbike with his right hand.\\
\textbf{Target Narration: }\#C C holds the part of the motorbike with his left hand.\\
\textbf{Instruction: }Person holds it with the other hand.\\\\
\textbf{Source Narration: }\#C c climbs up the steps.\\
\textbf{Target Narration: }\#C c climbs down the steps\\
\textbf{Instruction: }The person climbs down.\\\\
\textbf{Source Narration: }\#C C Wipes as paint brush with a paper towel.\\
\textbf{Target Narration: }\#C C dips brush in water.\\
\textbf{Instruction: }Dip the object in water.\\\\
\textbf{Source Narration: }\#C C pours the water in the shoe.\\
\textbf{Target Narration: }\#C C rinses the shoe.\\
\textbf{Instruction: }Rinse it instead.\\\\
\textbf{Source Narration: }\#C C puts electric shoe cleaner on the sink\\
\textbf{Target Narration: }\#C C puts shoe in the sink\\
\textbf{Instruction: }Same action with a shoe.\\\\
\textbf{Source Narration: }\#C C puts down penetrant oil\\
\textbf{Target Narration: }\#C C sprays the oil\\
\textbf{Instruction: }Spray it.\\\\
\textbf{Source Narration: }\#C C moves the scissors aside\\
\textbf{Target Narration: }\#C C moves the coins aside\\
\textbf{Instruction: }Change it to coins.\\\\
\textbf{Source Narration: }\#C C fixes the lawn mower basket\\                                 
\textbf{Target Narration: }\#C C holds the lawn mower basket\\
\textbf{Instruction: }Hold it instead.\\\\
\textbf{Source Narration: }\#c c sits on the mat\\
\textbf{Target Narration: }\#C C kneels on the carpet\\
\textbf{Instruction: }Kneels instead.\\\\
\textbf{Source Narration: }\#C C drops the plate of food on the sink slap.\\
\textbf{Target Narration: }\#C C picks a plate from the sink slap.\\
\textbf{Instruction: }Pick it up.\\\\
\textbf{Source Narration: }\#C C holds the basket of flowers on the floor with her left hand.\\
\textbf{Target Narration: }\#C C puts the white flowers on the tray with her right hand.\\
\textbf{Instruction: }Transfer it to the tray.\\\\
\textbf{Source Narration: }\#C C scrapes carrot remains from the grater into the brown bowl\\
\textbf{Target Narration: }\#C C picks carrot from the brown bowl with her right hand\\
\textbf{Instruction: }Pick it up from the bowl.
\end{tcolorbox}

\subsection{Creating Target Caption}\label{target} 
We present the prompt utilized to obtain a valid target caption from the query video caption and the instruction text. Our goal is to take a video caption along with an instruction specifying some changes and then generate a valid target caption. Similar to the dataset generation process, this also uses a few in-context examples to improve the quality of the generated captions. 

\begin{tcolorbox}[
enhanced,
colback=white!98!gray,
colframe=black, 
boxrule=.5pt,
colbacktitle=white!88!gray,
coltitle=black, 
fonttitle=\bfseries,
title={TF-CVR Prompt}, 
breakable=true,
]
\scriptsize
I have a video. Given a brief description of the source video and a instruction that modifies it, write a description of the target video. Keep the modified description as short as possible, while being complete. Mention objects only when absolutely necessary. Use the examples below for reference.\\\\
\textbf{Source Narration:} \#C C picks up the jug.\\
\textbf{Instruction:} The person is cleaning.\\
\textbf{Target Narration:} \#C C cleans the jug.\\\\
\textbf{Source Narration:} \#C C picks a spanner from the table with his right hand.\\
\textbf{Instruction:} Gasket being picked up. \\
\textbf{Target Narration:} \#C C picks a gasket from a table with his right hand.\\\\
\textbf{Source Narration:} \#C C picks up the wood from the shelf with his left hand.\\
\textbf{Instruction:} Person uses the other hand and detaches.\\
\textbf{Target Narration:} \#C C detaches a wood from the wooden structure with his right hand.\\\\
\textbf{Source Narration:} \#C C fixes the bolt on the motorbike with his right hand.\\
\textbf{Instruction:} Person holds it with the other hand.\\
\textbf{Target Narration:} \#C C holds the part of the motorbike with his left hand.\\\\
\textbf{Source Narration:} \#C c climbs up the steps.\\
\textbf{Instruction:} The person climbs down.\\
\textbf{Target Narration:} \#C c climbs down the steps.\\\\
\textbf{Source Narration:} \#C C Wipes as paint brush with a paper towel.\\
\textbf{Instruction:} Dip the object in water. \\
\textbf{Target Narration:} \#C C dips brush in water.\\\\
\textbf{Source Narration:} \#C C pours the water in the shoe.\\
\textbf{Instruction:} Rinse it instead.\\
\textbf{Target Narration:} \#C C rinses the shoe.\\\\
\textbf{Source Narration:} \#C C puts electric shoe cleaner on the sink.\\
\textbf{Instruction:} Same action with a shoe.\\
\textbf{Target Narration:} \#C C puts shoe in the sink.\\\\
\textbf{Source Narration:} \#C C puts down penetrant oil.\\
\textbf{Instruction:} Spray it.\\
\textbf{Target Narration:} \#C C sprays the oil.\\\\
\textbf{Source Narration:} \#C C moves the scissors aside.\\
\textbf{Instruction:} Change it to coins.\\
\textbf{Target Narration:} \#C C moves the coins aside.\\\\
\textbf{Source Narration:} \#C C fixes the lawn mower basket.\\                  
\textbf{Instruction:} Hold it instead.\\
\textbf{Target Narration:} \#C C holds the lawn mower basket.\\\\
\textbf{Source Narration:} \#C C sits on the mat.\\
\textbf{Instruction:} Kneels instead.\\                                      
\textbf{Target Narration:} \#C C kneels on the carpet.\\\\
\textbf{Source Narration:} \#C C drops the plate of food on the sink slap.\\
\textbf{Instruction:} Pick it up.\\
\textbf{Target Narration:} \#C C picks a plate from the sink slap.\\\\
\textbf{Source Narration:} \#C C holds the basket of flowers on the floor with her left hand.\\
\textbf{Instruction:} Transfer it to the tray.\\
\textbf{Target Narration:} \#C C puts the white flowers on the tray with her right hand.\\\\
\textbf{Source Narration:} \#C C scrapes carrot remains from the grater into the brown bowl.\\
\textbf{Instruction:} Pick it up from the bowl.\\
\textbf{Target Narration:} \#C C picks carrot from the brown bowl with her right hand.
\end{tcolorbox}

\subsection{Analysing Modification Instructions for Temporal vs. Object Events}\label{temporal} 
We analyse video modification instructions for both WebVid-CoVR~\cite{ventura2023covr} and our EgoCVR benchmark to study the type of modifications existing in the data sets. To categorize modification instructions as temporal or object-centric, we employ GPT-4. In the prompt, apart from a description of the task itself, we also provide a few in-context examples shown below. 

\begin{tcolorbox}[
float=htb,
enhanced,
colback=white!98!gray,
colframe=black, 
boxrule=.5pt,
colbacktitle=white!88!gray,
coltitle=black, 
fonttitle=\bfseries,
title={Temporal Event Detection Prompt}, 
]
\scriptsize
I have an instruction to modify a video. Looking at just the instruction, you should decide whether the instruction is focused on temporal events such as actions, or if it is just focused on objects. The answer you generate should only be "yes" or "no". Use the examples below for reference.\\\\
\textbf{Instruction:} have him fishing\\
\textbf{Answer:} yes\\\\
\textbf{Instruction:} turn it red on a watercolor stain.\\
\textbf{Answer:} no\\\\
\textbf{Instruction:} make the sax player into a drummer\\
\textbf{Answer:} no\\\\
\textbf{Instruction:} the girl is crying\\
\textbf{Answer:} yes\\\\
\textbf{Instruction:} change the meat to prawns\\
\textbf{Answer:} no\\\\
\textbf{Instruction:} remove the man.\\
\textbf{Answer:} no\\\\
\textbf{Instruction:} dip it into the paint.\\
\textbf{Answer:} yes\\\\
\textbf{Instruction:} cut the carrot instead.\\
\textbf{Answer:} no\\\\
\textbf{Instruction:} insert it into the roof.\\
\textbf{Answer:} yes\\\\
\textbf{Instruction:} pick it up instead.\\
\textbf{Answer:} yes
\end{tcolorbox}

\FloatBarrier
\section{Additional Qualitative Examples}\label{illustrations}
We illustrate a few more examples extracted from \benchmarkName in Figure \ref{fig:dataset_qualitative}. It is clearly visible that the data set contains a wide variety of activities in different environments, while the examples typically focus on action focused changes. 

\begin{figure}[ht]
    \centering
    \includegraphics[width=1.0\textwidth]{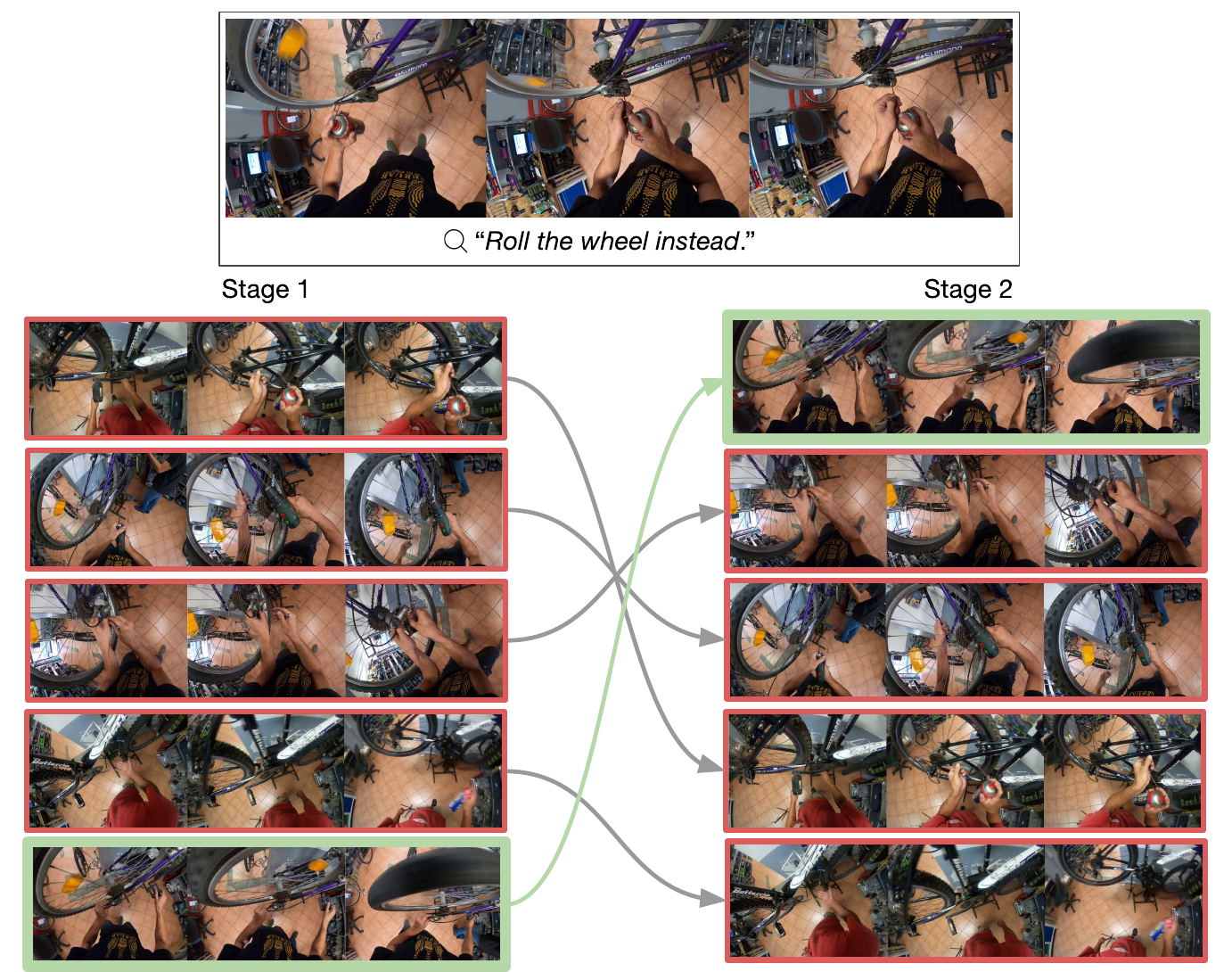}
    \caption{Qualitative example for the first and the second stage ranking results of our \methodName method for the query instruction ``\texttt{Roll the wheel instead.}''. The arrows indicate how the ranking was changed after re-ranking. The correct video is showcased in green.
    }
    \label{fig:stage1_vs_stage2}
\end{figure}

\begin{figure}[ht]
    \centering
    \includegraphics[width=1.0\textwidth]{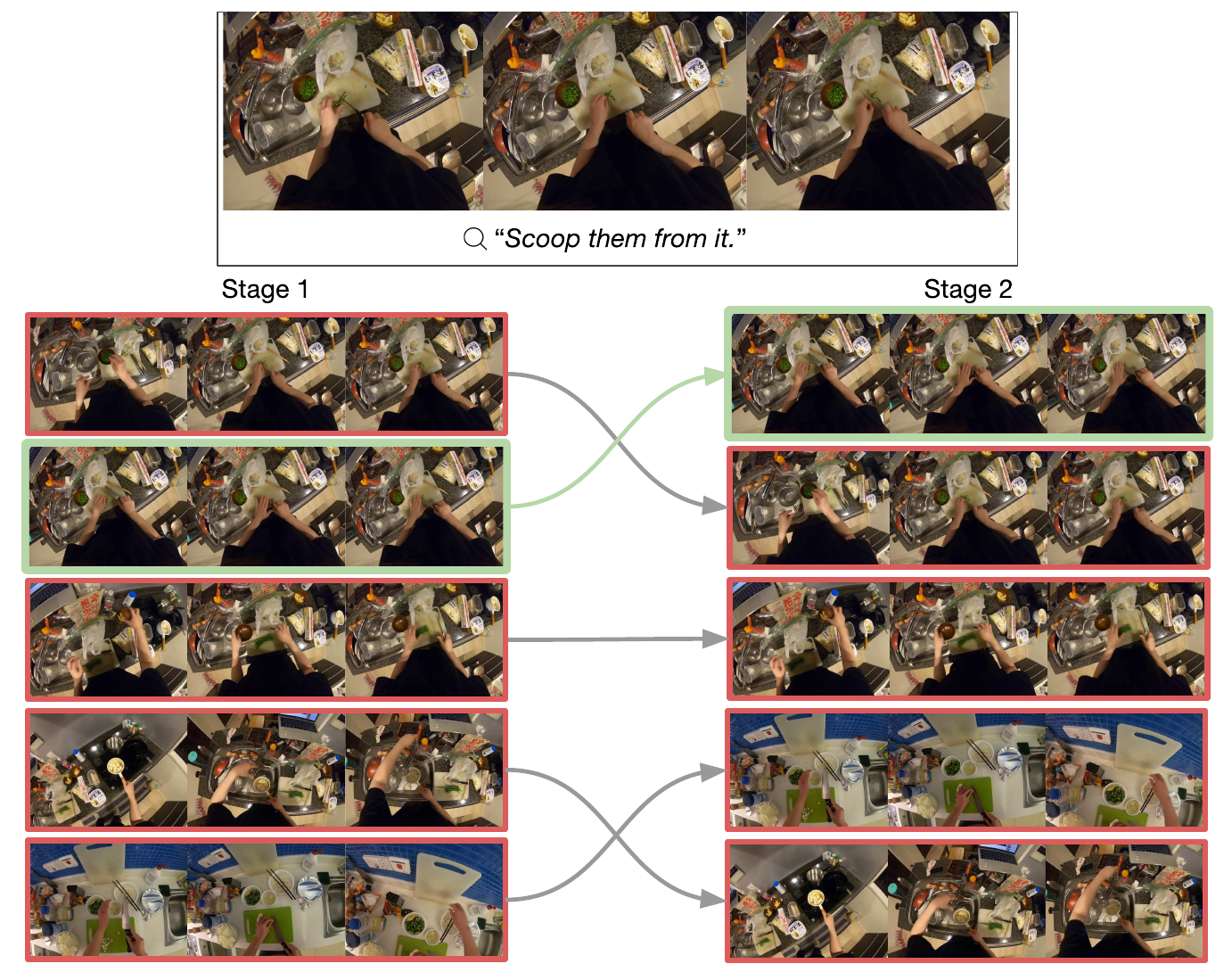}
    \caption{Qualitative example for the first and the second stage ranking results of our \methodName method for the query instruction ``\texttt{Scoop them from it.}''. The arrows indicate how the ranking was changed after re-ranking. The correct video is showcased in green.
    }
    \label{fig:stage1_vs_stage2_2}
\end{figure}

We also illustrate resulting order of the videos obtained after re-ranking performed by \methodName in Figure~\ref{fig:stage1_vs_stage2} and Figure~\ref{fig:stage1_vs_stage2_2}. We observe that the correct video is not in the first position in the first stage, however, after performing the second stage, the correct video is in the top position.

\begin{figure}[ht]
    \centering
    \includegraphics[width=1.0\textwidth]{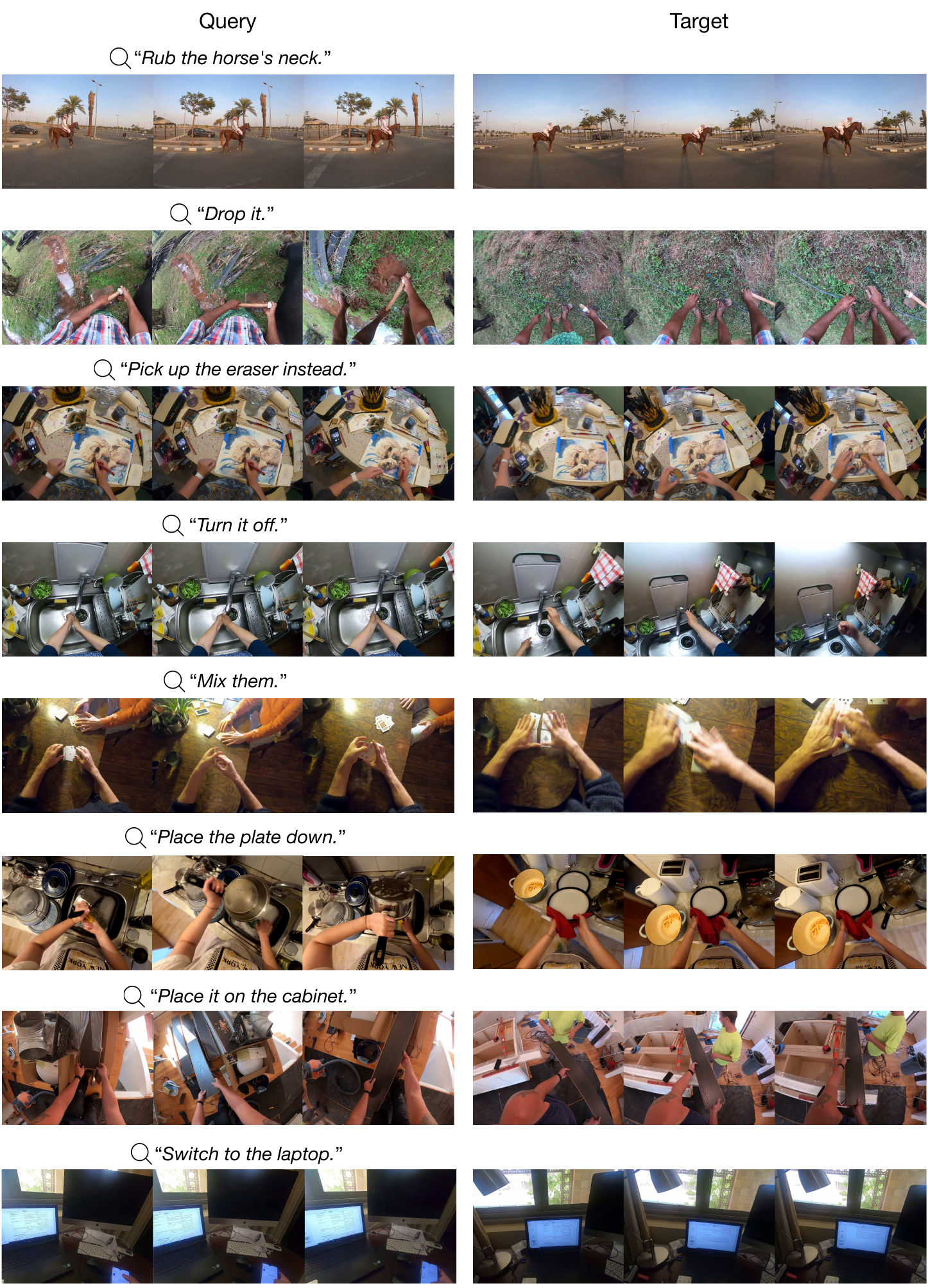}
    \caption{  
    Additional examples from our \benchmarkName benchmark. We present the input video, the text modification and the target video. The dataset is diverse, covering various types of scenes and activities.
    }
    \label{fig:dataset_qualitative}
\end{figure}

\FloatBarrier

\end{document}